\documentclass{article}

\usepackage{arxiv}

\usepackage[utf8]{inputenc} 
\usepackage[T1]{fontenc}    
\usepackage{hyperref}       
\usepackage{url}            
\usepackage{booktabs}       
\usepackage{amsfonts}       
\usepackage{nicefrac}       
\usepackage{microtype}      
\usepackage{lipsum}		
\usepackage{graphicx}

\usepackage{doi}
\usepackage{tikz}
\usepackage{pgfplots}
\pgfplotsset{compat=1.18}  

\usepackage[numbers]{natbib}
\usepackage{graphicx}
\usepackage{amsmath}
\usepackage{colortbl}
\usepackage{array}
\usepackage{booktabs}
\usepackage{colortbl}
\usepackage{xcolor}
\usepackage{multirow}
\usepackage{amssymb}
\usepackage{subcaption}
\usepackage{tabularray}
\usepackage{colortbl}
\usepackage{tabularx}
\usepackage{tikz-dependency}
\usepackage{subcaption}
\usepackage{float}

\usepackage{caption}
\usepackage{adjustbox}

\definecolor{aspectColor}{rgb}{0.6,0.8,1}        
\definecolor{positiveGreen}{rgb}{0.7,1,0.7}       
\definecolor{opinionColor}{rgb}{1,0.8,0.6}      
\definecolor{aspectClass}{rgb}{0.7,1,0.9}     
\definecolor{negativeRed}{rgb}{1,0.7,0.7}       

\title{AdaptiSent: Context-Aware Adaptive Attention for Multimodal Aspect-Based Sentiment Analysis}

\author{%
  \href{https://orcid.org/0000-0001-9404-1556}{\includegraphics[scale=0.06]{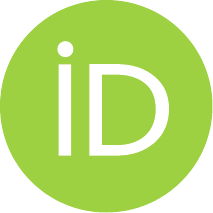}\hspace{1mm}S M Rafiuddin}\thanks{Corresponding Author}\\
  Department of Computer Science\\
  Oklahoma State University\\
  Stillwater, Oklahoma, USA\\
  \texttt{srafiud@okstate.edu}
  \And
  \href{https://orcid.org/0000-0003-0353-0894}{\includegraphics[scale=0.06]{orcid.pdf}\hspace{1mm}Sadia Kamal}\\
  Department of Computer Science\\
  Oklahoma State University\\
  Stillwater, Oklahoma, USA\\
  \texttt{sadia.kamal@okstate.edu}
  \And
  \href{https://orcid.org/0000-0001-6201-3729}{\includegraphics[scale=0.06]{orcid.pdf}\hspace{1mm}Mohammed Rakib}\\
  Department of Computer Science\\
  Oklahoma State University\\
  Stillwater, Oklahoma, USA\\
  \texttt{mohammed.rakib@okstate.edu}
  \And
  \href{https://orcid.org/0000-0002-7135-4602}{\includegraphics[scale=0.06]{orcid.pdf}\hspace{1mm}Arunkumar Bagavathi}\\
  Independent Researcher\\
  \texttt{b.arun410@gmail.com}
  \And
  \href{https://orcid.org/0000-0002-1640-5905}{\includegraphics[scale=0.06]{orcid.pdf}\hspace{1mm}Atriya Sen}\\
  Department of Computer Science\\
  Oklahoma State University\\
  Stillwater, Oklahoma, USA\\
  \texttt{atriya.sen@okstate.edu}
}

\renewcommand{\headeright}{ASONAM 2025}
\renewcommand{\undertitle}{Accepetd at ASONAM 2025}
\renewcommand{\shorttitle}{AdaptiSent: Context-Aware Adaptive Attention for Multimodal ABSA}

\hypersetup{
pdftitle={A template for the arxiv style},
pdfsubject={q-bio.NC, q-bio.QM},
pdfauthor={David S.~Hippocampus, Elias D.~Striatum},
pdfkeywords={First keyword, Second keyword, More},
}

\begin{document}
\maketitle

\begin{abstract}
We introduce AdaptiSent, a new framework for Multimodal Aspect-Based Sentiment Analysis (MABSA) that uses adaptive cross-modal attention mechanisms to improve sentiment classification and aspect term extraction from both text and images. Our model integrates dynamic modality weighting and context-adaptive attention, enhancing the extraction of sentiment and aspect-related information by focusing on how textual cues and visual context interact. We tested our approach against several baselines, including traditional text-based models and other multimodal methods. Results from standard Twitter datasets show that AdaptiSent surpasses existing models in precision, recall, and F1 score, and is particularly effective in identifying nuanced inter-modal relationships that are crucial for accurate sentiment and aspect term extraction. This effectiveness comes from the model's ability to adjust its focus dynamically based on the context's relevance, improving the depth and accuracy of sentiment analysis across various multimodal data sets. AdaptiSent sets a new standard for MABSA, significantly outperforming current methods, especially in understanding complex multimodal information.\footnote{For code and dataset, please contact: \textit{\textbf{srafiud@okstate.edu}}}
\end{abstract}



\section{Introduction}

The rise of social media has led to an abundance of multimodal content that blends text, images, and other media. While this enriches expression, it also complicates sentiment understanding—particularly when sentiments are tied to specific aspects. Multimodal Aspect-Based Sentiment Analysis (MABSA) addresses this challenge by jointly analyzing textual and visual signals to infer aspect-specific sentiment.

Historically, sentiment analysis mainly focused on text. The growth of multimodal data on social media required more advanced methods capable of interpreting the complex relationship between text and images. Significant developments in MABSA include the Cross-Modal Multitask Transformer by Yang \textit{et al.} (2022), which integrates visual data into text analysis, greatly improving performance \cite{yang2022cross}. Zhu \textit{et al.} (2015) have emphasized the importance of using linguistic structures in their research \cite{zhujoint}.

\definecolor{darkgreen}{rgb}{0,0.5,0}
\begin{table}[H]
\centering
\setlength{\tabcolsep}{4pt} 
\renewcommand{\arraystretch}{1.2} 
\caption{Multimodal sentiment examples with image, text, aspect term, and sentiment.}
\begin{tabular}{m{0.3\linewidth} m{0.3\linewidth} m{0.3\linewidth}}
\includegraphics[width=\linewidth]{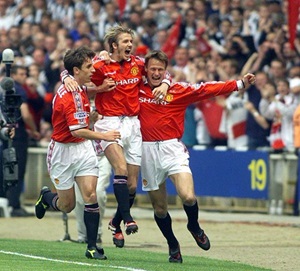} & \includegraphics[width=\linewidth]{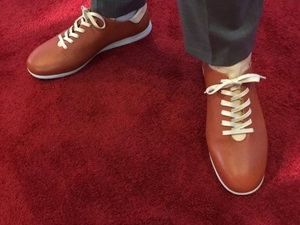} & \includegraphics[width=\linewidth]{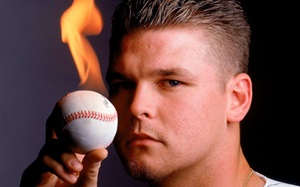} \\
Gary Neville, \textbf{\$T\$} and Teddy Sheringham celebrate for Manchester United. & Me listening to \textbf{\$T\$} sing \#BETAwards \#BETAwards17 & The media has lost all chill with \textbf{\$T\$}'s new documentary \\
\textit{\textbf{David Beckham}} & \textit{\textbf{Trey Songz}} & \textit{\textbf{Chris Brown}} \\
\textcolor{darkgreen}{\textbf{Positive}} & \textcolor{blue}{\textbf{Neutral}} & \textcolor{red}{\textbf{Negative}} \\
\end{tabular}
\end{table}

Recent advances leverage large pre-trained transformers and cross-modal attention to fuse text and image features for multimodal aspect-based sentiment analysis \cite{hu2024vision,fan2024position}. However, most methods apply direct fusion without addressing the modality gap—the differing ways text and images encode sentiment—which can lead to semantic inconsistencies and reduced performance \cite{xiang2023msfnet,xu2023multimodal}. While text often expresses opinions explicitly, images offer implicit emotional cues that may reinforce or contradict the sentiment \cite{yang2022face}. Many models either assume equal visual importance or ignore visual data when uncertain \cite{peng2024novel}. Though selective fusion and semantic-bridging strategies have emerged \cite{wang2024tmfn}, they often fail to capture fine-grained aspect alignment or adaptively weight multimodal signals.

This paper presents a new MABSA framework with five key features: \textbf{(1)} dynamic importance scoring to focus on relevant cues; \textbf{(2)} context‑aware weighting of text and images; \textbf{(3)} adaptive masking for each aspect; \textbf{(4)} aspect‑specific captioning with custom balancing; and \textbf{(5)} multimodal semantic alignment to integrate text and visual information. Unlike prior work that performs static fusion or treats visual inputs uniformly, \textbf{AdaptiSent} adaptively modulates attention weights based on per-aspect contextual importance, leveraging both learned linguistic and visual salience.

This study enhances sentiment analysis on social media by addressing challenges in semantic alignment and multimodal integration. It introduces the Enhanced Cross-Modal Attention Mechanism, followed by experiments on benchmark datasets. Results demonstrate improvements over prior models, with the conclusion summarizing key insights and future directions.

\section{Related Work}

Recent research in MABSA has focused on improving how text and image data are combined. Key developments include new models that adjust visual input to text, enhance the use of syntactic structures, and incorporate aesthetic evaluations for better cross-modal understanding. Significant contributions include the Cross-Modal Multitask Transformer by Yang et al. (2022) \cite{yang2022cross}, Atlantis by Xiao et al. (2024) \cite{xiao2024atlantis}, and syntactic adaptive models by Zhu et al. (2015) \cite{zhujoint}. Chauhan et al. (2023) also achieved top results with a new transformer model \cite{chauhan2023transformer}.

Attention to cross-modal interaction has led to methods that use facial expressions to improve text sentiment analysis \cite{yang2022face}, refine data integration \cite{feng2024autonomous}, and achieve nuanced data fusion \cite{wang2024tmfn,wang2024self}.

The role of pre-trained models and attention mechanisms has been explored to enhance the integration and alignment of multimodal data \cite{hu2024vision,fan2024position,xu2023multimodal}. Approaches like using external knowledge bases \cite{vargas2022europi}, addressing few-shot learning challenges \cite{yang2023few}, and syntax-aware hybrid prompting \cite{zhou2023syntax} have also been significant.

Despite progress, challenges in semantic alignment and noise reduction remain. Peng et al. (2024) introduce a novel energy-based model mechanism for multi-modal aspect-based sentiment analysis that explicitly models span pairwise relevance to improve visual–text alignment and achieves state-of-the-art performance on standard benchmarks. Innovative solutions like MSFNet \cite{xiang2023msfnet} and multi-curriculum denoising frameworks \cite{zhao2023m2df} are emerging to address these issues. Looking ahead, new machine learning techniques, such as energy-based models for enhancing visual-text relevance, are being explored \cite{peng2024novel}.

These advancements highlight a trend toward more sophisticated and effective MABSA models, leveraging both modalities' strengths to improve sentiment analysis applications.

\section{Method}

\subsection{Problem Formulation}

Multimodal Aspect-Based Sentiment Analysis (MABSA) aims to jointly extract aspect terms and predict their sentiments from a multimodal input comprising text \(\mathbf{T}^0 \in \mathbb{R}^{L \times d_t}\) and visual features \(\mathbf{V}_I \in \mathbb{R}^{K \times d_v}\), where \(L\) is the number of tokens, \(K\) the number of image regions or patches, and \(d_t, d_v\) are the respective embedding dimensions. Let \(\mathcal{A}\) denote the set of candidate aspect terms and \(\mathcal{S} = \{\texttt{positive}, \texttt{negative}, \texttt{neutral}\}\) the sentiment label space.

The goal is to identify a subset \(\mathcal{A}_{\mathrm{ext}}\subseteq\mathcal{A}\) and assign to each \(\mathbf{a}_i\in\mathcal{A}_{\mathrm{ext}}\) a sentiment \(\mathbf{s}_{\mathbf{a}_i}\in\mathcal{S}\), forming the output:
\begin{equation}\label{eq:aspect_sentiment}
\mathbf{D}
= \bigl\{\,(\mathbf{a}_i,\,\mathbf{s}_{\mathbf{a}_i}) \mid
\mathbf{a}_i\in\mathcal{A}_{\mathrm{ext}},\ 
\mathbf{s}_{\mathbf{a}_i} = \boldsymbol{f}(\mathbf{a}_i,\,\mathbf{T}^0,\,\mathbf{V}_I)\bigr\}.
\end{equation}
Here \(\boldsymbol{f}\colon \mathcal{A}\times\mathbb{R}^{L\times d_t}\times\mathbb{R}^{K\times d_v}\to\mathcal{S}\) is a multimodal sentiment classification function, and \(\mathbf{D}\subseteq\mathcal{A}\times\mathcal{S}\).

\subsection{Multimodal Representation}

\textbf{Textual Representation:} The text input is tokenized via RoBERTa's Byte-Pair Encoding into \( L \) tokens, including special tokens \( t_{\text{cls}} \) and \( t_{\text{sep}} \). Each token \( t_i \) is mapped to an embedding \( E(t_i) \in \mathbb{R}^{d_t} \), summed with positional encoding \( P_i \in \mathbb{R}^{d_t} \), yielding \( T^0 \in \mathbb{R}^{(L+2) \times d_t} \).

\textbf{Visual Representation:} The image \( I \) is divided into \( K \) patches, each projected to \( E(p_i) \in \mathbb{R}^{d_v} \) using a linear patch embedding. A special token \( p_{\text{cls}} \) is prepended, and positional embeddings \( P_i \in \mathbb{R}^{d_v} \) are added, resulting in \( V_I \in \mathbb{R}^{(K+1) \times d_v} \).\footnote{We denote several scalar parameters throughout the paper (e.g., \(\alpha_m\), \(\alpha_j\), \(\gamma\)). See Table~\ref{tab:parameters} for their definitions and values.}

\begin{figure*}[ht]
  \centering
  \includegraphics[width=1.0\textwidth]{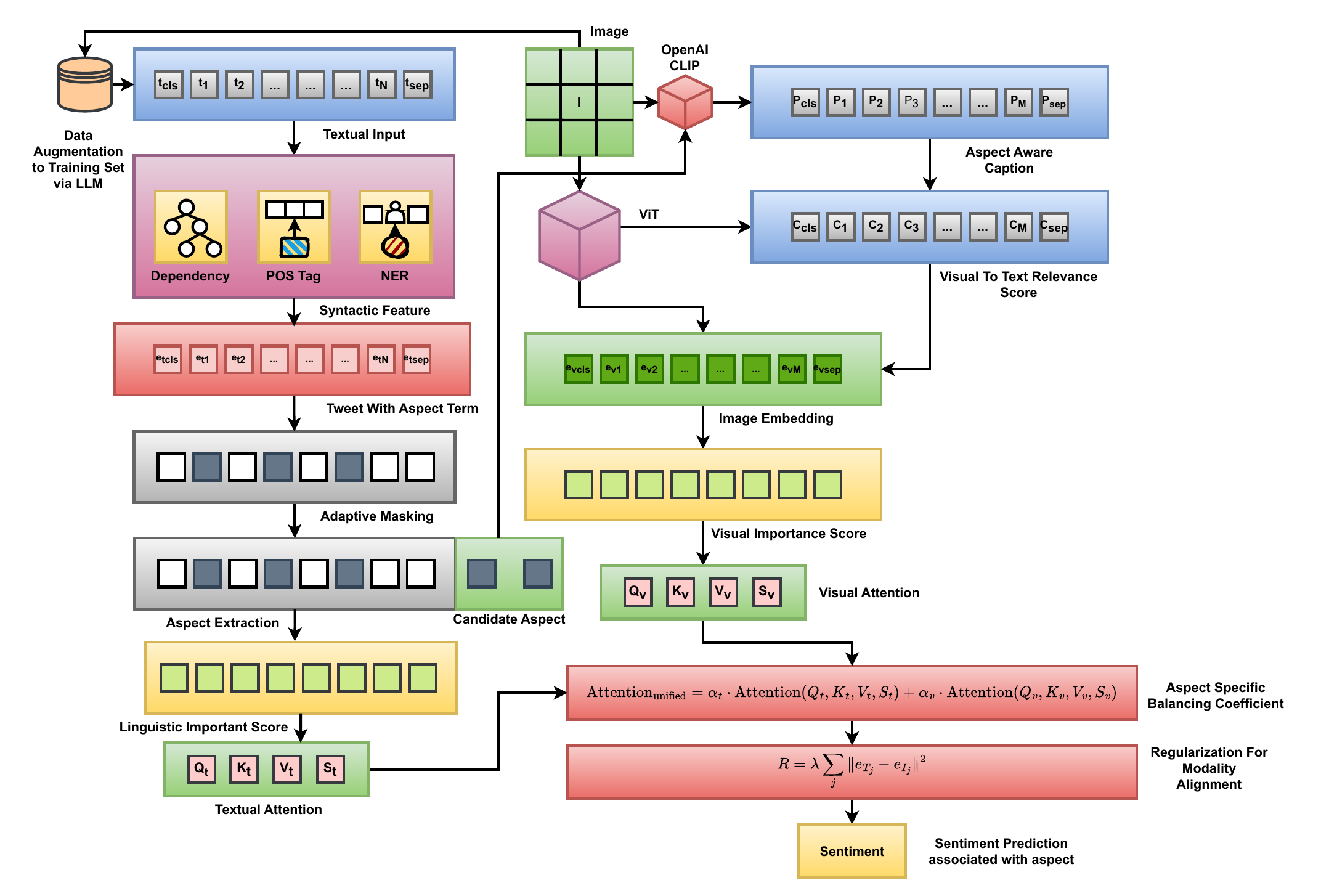}
  \caption{Overview of the \textbf{AdaptiSent} framework for MABSA. Given a tweet and its paired image, an \texttt{LLM} augments the input with aspect terms. Linguistic features (dependency, POS, NER) guide adaptive masking, and the masked text is encoded by \texttt{RoBERTa}. Simultaneously, \texttt{CLIP}~\cite{radford2021learning} generates aspect-aware captions and \texttt{ViT} extracts patch-level visual features. A visual-to-text relevance module assigns importance scores, fused via cross-modal self-attention modulated by aspect-specific coefficients. The final representation is regularized for modality alignment and used for per-aspect sentiment prediction.}
  \label{fig:mabsa}
\end{figure*}

 The inputs are embedded as \( T^0 \), \( V_I \), and \( C^0 \) respectively. Aspect-aware captions \( C^0 \) complement visual embeddings by providing additional semantic context that may not be fully captured by image features alone. Linguistic features—dependency trees \( D_T \), POS tags \( P_T \), and NER tags \( N_T \)—are also extracted to enrich the text representation.

Each token \( t_i \in T \) is mapped to a composite embedding:
\begin{equation}
\mathbf{e}_i = \mathbf{w}_i \oplus \mathbf{p}_i \oplus \mathbf{d}_i
\end{equation}
where \( \mathbf{w}_i \in \mathbb{R}^{d_t} \) is the word embedding, \( \mathbf{p}_i \in \mathbb{R}^{d_p} \) the POS embedding, and \( \mathbf{d}_i \in \mathbb{R}^{d_d} \) the dependency embedding. The fused features capture lexical, syntactic, and semantic information, supporting accurate aspect term extraction under multimodal context.

\subsection{Method for Multimodal Aspect Term Extraction: }

\subsubsection{Importance Score Computation}

\textbf{Visual-to-Text Relevance:} We compute visual relevance scores $\mathbf{R}_{\text{vis}}(t_i)$ by aggregating attention-based alignments between token embeddings $\mathbf{E}[t_i]$ and multimodal embeddings $\mathbf{V}_I, \mathbf{C}^0$ as:
\begin{equation}
\mathbf{R}_{\text{vis}}(t_i) = \mathrm{softmax}\bigl(\mathrm{att}(\mathbf{E}[t_i], \mathbf{V}_I) + \mathrm{att}(\mathbf{E}[t_i], \mathbf{C}^0)\bigr)
\end{equation}

\textbf{Linguistic Importance:} Linguistic importance scores $\mathbf{R}_{\text{ling}}(t_i)$ integrate dependency ($D_T$), POS ($P_T$), and NER ($N_T$) embeddings via a trainable linear combination:
\begin{equation}
\mathbf{R}_{\text{ling}}(t_i) = \mathrm{sigmoid}\bigl(\mathbf{W}_d\,\mathbf{d}_i + \mathbf{W}_p\,\mathbf{p}_i + \mathbf{W}_n\,\mathbf{n}_i + b\bigr)
\end{equation}
where $\mathbf{W}_d \in \mathbb{R}^{1\times d_d}$, $\mathbf{W}_p \in \mathbb{R}^{1\times d_p}$, $\mathbf{W}_n \in \mathbb{R}^{1\times d_n}$, and bias $b\in\mathbb{R}$ are learnable parameters optimized during training. Here, $\mathbf{d}_i$, $\mathbf{p}_i$, and $\mathbf{n}_i$ represent dependency, POS, and NER embeddings respectively. This parameterized approach allows the model to automatically learn the importance of each linguistic cue for optimal aspect extraction.
\textbf{Adaptive Masking:} Instead of a fixed threshold, an adaptive threshold $\theta$ is computed per sentence based on the variability of token importance scores $\mathbf{S}(t_i)$:
\begin{equation}
\theta = \mu_S + \alpha_m\,\sigma_S
\end{equation}
where $\mu_S$ and $\sigma_S$ are the mean and standard deviation of $\mathbf{S}(t_i)$, and $\alpha_m$ is a learnable scaling parameter specific to masking. Tokens are then masked as:
\begin{equation}
m(t_i) = 
\begin{cases}
\texttt{[MASK]} & \text{if } \mathbf{S}(t_i) > \theta, \\
t_i & \text{otherwise.}
\end{cases}
\end{equation}

\textbf{Aspect Term Prediction:} The masked sequence $m(T^0)$ is fed into a \texttt{RoBERTa}-based extractor, augmented with visual features $\mathbf{V}_I$ and aspect-aware captions $\mathbf{C}^0$, to predict extracted aspects:
\begin{equation}
\mathcal{A}_{\text{ext}} = \texttt{RoBERTa}_{\text{masked}}\bigl(m(T^0), \mathbf{V}_I, \mathbf{C}^0\bigr)
\end{equation}
\texttt{RoBERTa} classifies each token, leveraging multimodal context to identify aspect terms.

\subsection{Method for Multimodal Aspect based Sentiment Classification: }

\subsubsection{Visual-Guided Textual Data Augmentation}

To enhance multimodal training diversity, we propose a visual-guided textual data augmentation strategy. Given an original text $\mathbf{T}$, associated image $I$, and extracted candidate aspects $\mathcal{A}_{\text{ext}}$, the image is first encoded into an embedding $\mathbf{e}_I \in \mathbb{R}^{d}$ via a pre-trained \texttt{ViT}. Large language model (\texttt{GPT 3.5 and Llama 3.0}) then generates augmented text $\mathbf{T}'$, conditioned on the original text, visual embedding, and candidate aspects:
\begin{equation}
\mathbf{T}' = \texttt{LLM}_{\text{aug}}\bigl(\mathbf{T}, \mathbf{e}_I, \mathcal{A}_{\text{ext}}\bigr)
\end{equation}

The augmented text $\mathbf{T}'$ is encoded using \texttt{RoBERTa}, producing a textual embedding $\mathbf{e}_{T'_j} \in \mathbb{R}^{d}$ consistent with the original textual encoding. To ensure alignment between the augmented text and visual content, we calculate their coherence via cosine similarity:
\begin{equation}
\text{Coherence}\bigl(\mathbf{e}_{T'_j}, \mathbf{e}_I\bigr) = \frac{\mathbf{e}_{T'_j} \cdot \mathbf{e}_I}{\|\mathbf{e}_{T'_j}\|\|\mathbf{e}_I\|}
\end{equation}

The augmented textual embeddings $\mathbf{e}_{T'_j}$, along with original textual and visual embeddings, are incorporated into the training set. This enrichment improves the model's ability to effectively interpret multimodal inputs, ultimately enhancing performance on multimodal aspect-based sentiment classification tasks.

\subsubsection{Aspect-Specific Balancing Coefficients}

To adaptively control the contribution of text and image modalities for each aspect term $a_j$, we introduce a learnable balancing coefficient $\alpha_j$. This allows the model to dynamically emphasize either textual or visual features based on contextual relevance during sentiment classification.

The textual embedding $\mathbf{e}_{T_j}$ is extracted using a \texttt{RoBERTa} encoder conditioned on the input text $\mathbf{T}$ and candidate aspects $\mathcal{A}_{\text{ext}}$, while the visual embedding $\mathbf{e}_{I_j}$ is obtained via a \texttt{ViT} processing the associated image $I$ and aspect-aware caption $C$.

The fused representation for each aspect is computed by weighting $\mathbf{e}_{T_j}$ and $\mathbf{e}_{I_j}$ according to $\alpha_j$, where $\alpha_j$ is initialized uniformly (i.e., 0.5) and optimized through backpropagation alongside other model parameters.

\subsubsection{Context-Adaptive Cross-Modal Attention Mechanism}

We propose a cross-modal attention mechanism that dynamically integrates visual-to-text relevance and linguistic importance scores to enhance aspect-based sentiment analysis.

Given token-level linguistic $R_{\text{ling}}(t_i)$ and visual $R_{\text{vis}}(t_i)$ importance scores, we compute a combined importance score:
\begin{equation}
\mathbf{S}(t_i) = \gamma\, R_{\text{ling}}(t_i) + (1-\gamma)\, R_{\text{vis}}(t_i)
\label{eq:lingvis}
\end{equation}
where $\gamma \in [0,1]$ is a hyperparameter controlling the trade-off between linguistic and visual importance.

The standard scaled dot-product attention is modified to incorporate $\mathbf{S}$ as an adaptive bias:
\begin{equation}
\texttt{Attention}(Q, K, V, \mathbf{S}) = \text{softmax}\!\Bigl(\frac{QK^\top}{\sqrt{d_k}} + \beta\,\mathbf{S}\Bigr)\,V
\end{equation}

where $\beta$ is a trainable scaling factor learned during training.

To further adapt modality contributions, we compute modality weighting coefficients:
\begin{align}
\alpha_t &= \frac{\sum_{i} R_{\text{ling}}(t_i)}{\sum_{i} R_{\text{ling}}(t_i) + \sum_{i} R_{\text{vis}}(t_i)} \\
\alpha_v &= 1 - \alpha_t
\end{align}
assigning higher weights to the more informative modality.

The unified attention output combines modality-specific attentions:
\begin{equation}
\texttt{Attention}_{\text{unified}} = \alpha_t\,\texttt{Attention}(Q_t, K_t, V_t, \mathbf{S}_t) + \alpha_v\,\texttt{Attention}(Q_v, K_v, V_v, \mathbf{S}_v)
\end{equation}

allowing the model to dynamically focus on the most relevant cross-modal features.

Although additional computations are introduced through importance-based modulation, the context-adaptive attention remains efficient as it operates over token-level importance scores and only lightly modifies the standard attention mechanism without increasing the number of attention heads or layers, thus ensuring practical scalability during training.

\subsubsection{Regularization for Modality Alignment}

To encourage consistency between textual and visual embeddings for each aspect $a_j$, we introduce a regularization term.

Original embeddings from \texttt{RoBERTa} ($\mathbf{e}_{T_j} \in \mathbb{R}^{d_t}$) and \texttt{ViT} ($\mathbf{e}_{I_j} \in \mathbb{R}^{d_v}$) are first mapped via modality-specific linear projections into a common embedding space $\mathbb{R}^{d}$ to ensure dimensional compatibility:
\begin{equation}
\mathbf{e}_{T_j}' = \mathbf{W}_T\,\mathbf{e}_{T_j} + b_T,\quad
\mathbf{e}_{I_j}' = \mathbf{W}_I\,\mathbf{e}_{I_j} + b_I
\end{equation}
where $\mathbf{W}_T \in \mathbb{R}^{d \times d_t}$, $b_T \in \mathbb{R}^{d}$, $\mathbf{W}_I \in \mathbb{R}^{d \times d_v}$, and $b_I \in \mathbb{R}^{d}$ are trainable parameters.

The modality alignment distance is computed in the shared space using squared Euclidean distance:
\begin{equation}
d\bigl(\mathbf{e}_{T_j}', \mathbf{e}_{I_j}'\bigr) = \|\mathbf{e}_{T_j}' - \mathbf{e}_{I_j}'\|^2
\end{equation}

The regularization loss aggregates these distances across all aspects:
\begin{equation}
R = \lambda \sum_{j=1}^{m} \|\mathbf{e}_{T_j}' - \mathbf{e}_{I_j}'\|^2
\end{equation}
where $\lambda$ is a hyperparameter tuned via validation, controlling the strength of modality alignment during training.

\begin{table}[h]
\centering
\caption{Summary of key parameters and their selected values. Here, \(\gamma \in [0, 1]\) is a hyperparameter balancing linguistic and visual importance (see also Eq.~\ref{eq:lingvis}).}
\label{tab:parameters}
\begin{tabular}{|c|c|c|c|}
\hline
\textbf{Parameter} & \textbf{Role} & \textbf{Type} & \textbf{Value} \\
\hline
\(\boldsymbol{\alpha}_m\) & Masking threshold scaling & Trainable & — \\
\(\boldsymbol{\alpha}_j\) & Modality balancing coefficient & Trainable & — \\
\(\boldsymbol{\beta}\)     & Attention scaling factor    & Trainable & — \\
\(\boldsymbol{\gamma}\)    & Linguistic–visual balance   & Hyperparameter & 0.3 \\
\(\boldsymbol{\lambda}\)   & Modality alignment strength & Hyperparameter & 0.1 \\
\hline
\end{tabular}
\end{table}

\subsection{Training Procedure}

\subsubsection{Loss Function for MABSA}

The overall loss jointly optimizes aspect term extraction, sentiment classification, and modality alignment:
\begin{equation}
\boldsymbol{L} = \sum_{i=1}^{n} \boldsymbol{w}_i \cdot \text{CrossEntropy}\bigl(\boldsymbol{p}_i, y_i\bigr)
  + \boldsymbol{\lambda} \sum_{j=1}^{m} \bigl\|\mathbf{e}_{T_j}' - \mathbf{e}_{I_j}'\bigr\|^2
\end{equation}
Here, $\boldsymbol{p}_i$ is the predicted distribution for token $t_i$, $y_i$ is the ground-truth label, and $\boldsymbol{w}_i$ is a token-specific weight derived from visual $R_{\text{vis}}(t_i)$ and linguistic $R_{\text{ling}}(t_i)$ scores, modulated by trainable parameters $\boldsymbol{\alpha}_m$ (masking) and $\boldsymbol{\beta}$ (attention scaling).

The second term encourages alignment between projected text and image embeddings $\mathbf{e}_{T_j}'$, $\mathbf{e}_{I_j}' \in \mathbb{R}^{d}$, computed via trainable linear layers. The regularization strength $\boldsymbol{\lambda}$ is tuned through validation experiments. Modality balancing coefficients $\boldsymbol{\alpha}_j$ are trainable, while the fusion weight $\boldsymbol{\gamma}$ is a fixed hyperparameter controlling the linguistic–visual importance trade-off.

\section{Experiments}
\subsection{Datasets}
We evaluate our method on two widely-used Multimodal Aspect-Based Sentiment Analysis (MABSA) datasets: \textbf{Twitter-15} and \textbf{Twitter-17}, each containing tweets with paired text and images. An aspect prediction is considered correct only if both the extracted aspect term and its associated sentiment polarity match the ground truth. Dataset statistics are summarized in Table~\ref{tab:twitter_stats}, where \textbf{Aspects} denotes the average number of aspects per sample and \textbf{Length} refers to the average number of tokens per text.

\begin{table}[ht]
\centering
\caption{Statistics of Twitter-15 and Twitter-17 datasets. "Aspects" = avg. number of aspects per sample, "Length" = avg. tokens per text.}
\label{tab:twitter_stats}

\setlength{\tabcolsep}{3pt}
\begin{tabularx}{\columnwidth}{lXXXXXXX}
\toprule
\textbf{Twitter-15} & \textbf{Pos} & \textbf{Neg} & \textbf{Neu} & \textbf{Total} & \textbf{Aspects} & \textbf{Words} & \textbf{Length} \\ \midrule
\textbf{Train} & 928  & 368  & 1883 & 3179 & 1.348 & 9023 & 16.72 \\ 
\textbf{Dev}   & 303  & 149  & 670  & 1122 & 1.336 & 4238 & 16.74 \\ 
\textbf{Test}  & 317  & 113  & 607  & 1037 & 1.345 & 3919 & 17.05 \\ 
\bottomrule
\end{tabularx}

\vspace{0.4cm}

\begin{tabularx}{\columnwidth}{lXXXXXXX}
\toprule
\textbf{Twitter-17} & \textbf{Pos} & \textbf{Neg} & \textbf{Neu} & \textbf{Total} & \textbf{Aspects} & \textbf{Words} & \textbf{Length} \\ \midrule
\textbf{Train} & 1508 & 1638 & 416  & 3562 & 1.410 & 6027 & 16.21 \\ 
\textbf{Dev}   & 515  & 517  & 144  & 1176 & 1.439 & 2922 & 16.37 \\ 
\textbf{Test}  & 493  & 573  & 168  & 1234 & 1.450 & 3013 & 16.38 \\ 
\bottomrule
\end{tabularx}
\end{table}

\subsection{Experimental Setup}

Experiments use pretrained \texttt{RoBERTa-base} \cite{liu2019roberta} and \texttt{ViT-base-patch16-224-in21k} weights to initialize our text and vision models. \texttt{RoBERTa} improves on \texttt{BERT} by using dynamic masking and larger training data. \texttt{ViT} splits each image into $16\times16$ patches and applies self-attention over those patches, making it well suited for vision tasks \cite{dosovitskiy2020image}.

Our models have a hidden size of $d = 768$, with 8 heads for cross‑modal self‑attention. \texttt{ViT} uses 16×16 pixel patches, matching the \texttt{ViT-base-patch16-224} configuration. We adopt the AdamW optimizer \cite{loshchilov2017decoupled} with a $2 \times 10^{-5}$ learning rate, incorporating a warmup phase. Settings include a 60‑token text length limit and batch size of 16. Experiments run on NVIDIA A100 GPUs with 24GB VRAM in PyTorch 1.9, generally concluding within 3 hours based on task complexity.

\section{Results}

\subsection{Compared Baseline Models}

\begin{table*}[htbp]
\centering
\caption{Performance comparison on MABSA datasets (\textbf{Twitter15} and \textbf{Twitter17}) with Precision (Prec), Recall (Rec), and F1 scores. Values in parentheses indicate standard deviation over 3 runs with different random seeds.}

\label{tab:table1}

\begin{tabular*}{\textwidth}{@{\extracolsep{\fill}}l ccc ccc}
\toprule
\textbf{Model} 
  & \multicolumn{3}{c}{\textbf{Twitter15}} 
  & \multicolumn{3}{c}{\textbf{Twitter17}} \\
\cmidrule(lr){2-4} \cmidrule(lr){5-7}
  & \textbf{Prec} & \textbf{Rec} & \textbf{F1} 
  & \textbf{Prec} & \textbf{Rec} & \textbf{F1} \\
\midrule
\multicolumn{7}{c}{\textbf{Text-Only Models}} \\
\midrule
SPAN \cite{hu2019open}        & 53.7 & 53.9 & 53.8  & 59.6 & 61.7 & 60.6  \\
D‑GCN \cite{chen2020joint}    & 58.3 & 58.8 & 58.6  & 64.2 & 64.1 & 64.1  \\
BART \cite{lewis2019bart}     & 62.9 & 65.0 & 63.9  & 65.2 & 65.6 & 65.4  \\
RoBERTa \cite{liu2019roberta} & 62.9 & 63.7 & 63.3  & 65.1 & 66.2 & 65.7  \\
\midrule
\multicolumn{7}{c}{\textbf{Multimodal Models}} \\
\midrule
UMT \cite{yu2020improving}      & 58.4 & 61.4 & 59.9  & 62.3 & 62.4 & 62.4  \\
OSCGA \cite{wu2020multimodal}   & 61.7 & 63.4 & 62.5  & 63.4 & 64.0 & 63.7  \\
JML \cite{ju2021joint}          & 65.0 & 63.2 & 64.1  & 66.5 & 65.5 & 66.0  \\
VLP \cite{ling2022vision}       & 68.3 & 66.6 & 67.4  & 69.2 & 68.0 & 68.6  \\
CMMT \cite{yang2022cross}       & 64.6 & 68.7 & 66.6  & 67.6 & 69.4 & 68.5  \\
M2DF \cite{zhao2023m2df}        & 67.0 & 68.3 & 67.6  & 67.9 & 68.8 & 68.4  \\
DTCA \cite{yu2022dual}          & 67.3 & 69.5 & 68.4  & 69.6 & 71.2 & 70.4  \\
AoM \cite{zhou2023aom}          & 67.9 & 69.3 & 68.6  & 68.4 & 71.0 & 69.7  \\
TMFN \cite{wang2024tmfn}        & 68.4 & 69.6 & 69.0  & 70.7 & 71.2 & 71.0  \\
DQPSA \cite{peng2024novel}                    & 71.7 & 72.0 & 71.9  & 71.1 & 70.2 & 70.6  \\
\midrule
\multicolumn{7}{c}{\textbf{Large Language Models}} \\
\midrule
Llama2 \cite{touvron2023llama}                  & 53.6 & 55.0 & 54.3  & 57.6 & 58.8 & 58.2  \\
Llama3 \cite{touvron2023llama}                  & 56.4 & 57.2 & 56.8  & 61.8 & 62.5 & 62.2  \\
GPT‑2.0 \cite{chatgpt}                          & 47.8 & 49.2 & 48.5  & 52.0 & 53.9 & 52.9  \\
GPT‑3.5 \cite{chatgpt}                          & 50.9 & 51.9 & 51.4  & 55.6 & 56.1 & 55.9  \\
\midrule
\textbf{AdaptiSent} & 70.9\,{\scriptsize(±0.27)} & 72.8\,{\scriptsize(±0.39)} & \textbf{71.9}\,{\scriptsize(±0.18)} & 71.4\,{\scriptsize(±0.52)} & 71.8\,{\scriptsize(±0.31)} & \textbf{71.6}\,{\scriptsize(±0.24)} \\

\bottomrule
\end{tabular*}
\end{table*}

\textbf{SPAN} \cite{hu2019open} introduces a span‑based extraction mechanism to resolve sentiment inconsistencies in text‑only settings, outperforming traditional sequence‑tagging methods by flexibly identifying sentiment spans. \textbf{D‑GCN} \cite{chen2020joint} incorporates syntactic dependencies via directional graph convolutions, yielding more precise joint aspect–sentiment representations. \textbf{BART} \cite{lewis2019bart} leverages denoising sequence to sequence pre‑training for robust text comprehension and implicit sentiment handling, while \textbf{RoBERTa} \cite{liu2019roberta} refines BERT’s training objectives and data scale to further enhance contextual understanding. 

Among multimodal approaches, \textbf{UMT} \cite{yu2020improving} unifies textual and visual encoders to inject visual context into sentiment inference, and \textbf{OSCGA} \cite{wu2020multimodal} employs dense co‑attention at both object and character granularities. \textbf{JML} \cite{ju2021joint}, \textbf{VLP} \cite{ling2022vision}, and \textbf{CMMT} \cite{yang2022cross} build on vision–language pre‑training with adaptive visual weighting, effectively balancing modalities. \textbf{M2DF} \cite{zhao2023m2df} and \textbf{DTCA} \cite{yu2022dual} exploit advanced transformer architectures and denoising channels to strengthen text–image synergy. \textbf{AoM} \cite{zhou2023aom} selectively aligns image regions to textual aspects, reducing noise in fusion, while \textbf{TMFN} \cite{wang2024tmfn} introduces multi‑grained feature fusion and target‑oriented alignment to emphasize emotion‑relevant cues. \textbf{DQPSA} \cite{peng2024novel} further refines cross‑modal gating and attention regularization to sharpen multimodal interactions. 

General‑purpose LLMs such as \textbf{Llama2}, \textbf{Llama3} \cite{touvron2023llama}, \textbf{GPT‑2.0} and \textbf{GPT‑3.5} \cite{chatgpt} exhibit strong language understanding but lack dedicated multimodal training, resulting in lower effectiveness on MABSA tasks. In contrast, \textbf{AdaptiSent} (Ours) combines LLM‑augmented aspect term insertion, syntactic‑guided masking, and learnable cross‑modal self‑attention—constrained by a modality‑alignment regularizer—to isolate genuine sentiment signals and set a new state‑of‑the‑art in multimodal aspect‑based sentiment analysis.

\subsection{Ablation Studies}

\begin{table*}[!h]
\centering

\caption{Ablation study for MABSA with different feature combinations, evaluated on \textbf{Twitter15} and \textbf{Twitter17}. Results are averaged over 3 runs with random seeds.}
\label{tab:table7}

\begin{tabularx}{\textwidth}{lcccccccc}
\toprule
\textbf{Model} & \multicolumn{4}{c}{\textbf{Twitter15}} & \multicolumn{4}{c}{\textbf{Twitter17}} \\
\cmidrule(lr){2-5} \cmidrule(lr){6-9}
 & \textbf{Acc} & \textbf{Prec} & \textbf{Rec} & \textbf{F1} 
 & \textbf{Acc} & \textbf{Prec} & \textbf{Rec} & \textbf{F1} \\
\midrule
w/o Aspect-Aware Captions                 & 72.33 & 67.13 & 63.51 & 65.27 & 73.17 & 68.37 & 65.53 & 66.92 \\
w/o Regularization for Modality Alignment & 73.58 & 67.89 & 64.44 & 66.12 & 77.71 & 70.22 & 66.26 & 68.18 \\
w/o Aspect-Specific Balancing Coefficients& 71.84 & 65.11 & 64.30 & 64.70 & 72.83 & 67.08 & 64.41 & 65.72 \\
w/o Data Augmentation                     & 76.85 & 74.56 & 66.64 & 70.38 & 78.94 & 74.50     & 67.68  & 70.93\\
w/o Context-Based Masking                 & 74.38 & 70.11 & 64.56 & 67.22 & 78.66 & 72.34 & 67.77 & 69.98 \\
\midrule
\textbf{AdaptiSent (Full Model)}          & \textbf{78.57} & \textbf{70.95} & \textbf{72.85} & \textbf{71.89}
                                          & \textbf{80.30} & \textbf{71.42} & \textbf{71.83} & \textbf{71.62} \\
\bottomrule
\end{tabularx}
\end{table*}

Table~\ref{tab:table7} shows each component’s impact. Removing aspect-specific balancing coefficients causes the largest F1 drop (71.89 → 64.70 on Twitter-15, 71.62 → 65.72 on Twitter-17; –7.19 pts and –5.90 pts), highlighting the need for adaptive modality weighting. Dropping aspect-aware captions is next (–6.62 pts, –4.70 pts), while the alignment regularizer and context masking provide moderate gains (–5.77 pts, –3.44 pts; –4.67 pts, –1.64 pts). Data augmentation has minimal effect (–1.51 pts, –0.69 pts). Figure~\ref{fig:hyperparam_sensitivity} shows our hyperparameters (\(\gamma=0.3\), \(\lambda=0.1\)) lie near the peaks. Together, these results confirm that each component contributes uniquely, with the full model achieving F1 scores of 71.89 on Twitter-15 and 71.62 on Twitter-17. This systematic analysis underscores the robustness of our design across both datasets.

\begin{figure}[!h]
  \centering
  \begin{subfigure}[b]{0.48\textwidth}
    \centering
    \begin{tikzpicture}
      \begin{axis}[
        width=\linewidth,
        height=\linewidth,
        every axis plot/.append style={line width=1.2pt},
        title={F1 vs.\ $\gamma$ (Linguistic–Visual Balance)},
        xlabel={$\gamma$}, ylabel={F1 Score (\%)},
        symbolic x coords={0.01,0.03,0.1,0.3,0.5,1},
        xtick=data,
        tick label style={font=\tiny},
        grid=both, grid style={gray!30,dashed},
        legend style={
          font=\tiny, draw=none,
          inner sep=1pt, outer sep=0pt,
          at={(0.98,0.98)}, anchor=north east,
          nodes={scale=0.6}
        },
        legend image post style={scale=0.6},
      ]
        \addplot[orange, mark=o, mark options={fill=white}]
          coordinates {
            (0.01,68.0) (0.03,69.5) (0.1,71.6) (0.3,71.7)
            (0.5,71.0)  (1,70.0)
          }; \addlegendentry{Twitter-15}
        \addplot[blue, mark=x, mark options={solid}]
          coordinates {
            (0.01,68.5) (0.03,69.8) (0.1,71.2) (0.3,71.5)
            (0.5,71.0)  (1,69.5)
          }; \addlegendentry{Twitter-17}
        \addplot[gray,dashed] coordinates {(0.3,65) (0.3,72.5)};
        \node[gray,rotate=90,font=\scriptsize]
          at (axis cs:0.3,67) {Opt.\ $\gamma=0.3$};
      \end{axis}
    \end{tikzpicture}
    \caption{F1 sensitivity to $\gamma$.}
    \label{fig:gamma_sensitivity}
  \end{subfigure}%
  \hfill
  \begin{subfigure}[b]{0.48\textwidth}
    \centering
    \begin{tikzpicture}
      \begin{axis}[
        width=\linewidth,
        height=\linewidth,
        every axis plot/.append style={line width=1.2pt},
        title={F1 vs.\ $\lambda$ (Modality-Alignment Strength)},
        xlabel={$\lambda$}, ylabel={F1 Score (\%)},
        symbolic x coords={0.01,0.03,0.1,0.3,0.5,1},
        xtick=data,
        tick label style={font=\tiny},
        grid=both, grid style={gray!30,dashed},
        legend style={
          font=\tiny, draw=none,
          inner sep=1pt, outer sep=0pt,
          at={(0.98,0.98)}, anchor=north east,
          nodes={scale=0.6}
        },
        legend image post style={scale=0.6},
      ]
        \addplot[orange, mark=o, mark options={fill=white}]
          coordinates {
            (0.01,67.0) (0.03,70.0) (0.1,71.5) (0.3,70.8)
            (0.5,70.3)  (1,68.0)
          }; \addlegendentry{Twitter-15}
        \addplot[blue, mark=x, mark options={solid}]
          coordinates {
            (0.01,67.5) (0.03,70.5) (0.1,71.0) (0.3,70.0)
            (0.5,68.8)  (1,67.8)
          }; \addlegendentry{Twitter-17}
        \addplot[gray,dashed] coordinates {(0.1,64) (0.1,72.5)};
        \node[gray,rotate=90,font=\scriptsize]
          at (axis cs:0.1,66.5) {Opt.\ $\lambda=0.1$};
      \end{axis}
    \end{tikzpicture}
    \caption{F1 sensitivity to $\lambda$.}
    \label{fig:lambda_sensitivity}
  \end{subfigure}
  \vspace{-1ex}
  \caption{Hyperparameter sensitivity: (a) variation with $\gamma$, peaking at $0.3$; (b) variation with $\lambda$, peaking at $0.1$.}
  \label{fig:hyperparam_sensitivity}
\end{figure}

\subsection{Case Studies}

\definecolor{deepgreen}{rgb}{0.1, 0.35, 0.1} 

\newcommand{\customcross}{%
  \tikz [x=1.5ex,y=1.5ex,line width=.15ex] \draw (0,0) -- (1,1) (0,1) -- (1,0);%
}

\newcommand{\textcross}{\ding{55}} 

\begin{table}[!h]
\centering

\setlength{\tabcolsep}{1pt}
\renewcommand{\arraystretch}{1.1}
\caption{Comparison of sentiment analysis models.}
\label{tab:table5}
\resizebox{\linewidth}{!}{%
\begin{tabular}{|m{1.5cm}|m{3.0cm}|m{2.2cm}|m{2.5cm}|m{2.5cm}|m{2.5cm}|m{2.5cm}|}
\hline
\textbf{Image} & \textbf{Text} & \textbf{Ground Truth} & \textbf{TMFN Model} & \textbf{AoM Model} & \textbf{DPQSA Model} & \textbf{Ours} \\
\hline
\includegraphics[width=1.5cm,height=1.5cm]{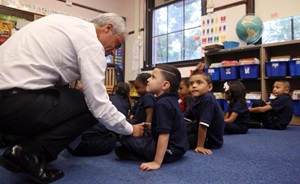} &
First day of school in Chicago and at Cameron Elementary. This kindergartener wasn’t impressed by the mayoral visit &
\textcolor{blue}{\textbf{(Chicago, Neutral)}} \newline \textcolor{red}{\textbf{(Cameron Elementary, Negative)}} &
\textcolor{blue}{\textbf{\checkmark\,(Chicago, Neutral)}} \newline \textcolor{deepgreen}{\textbf{\customcross\,(Cameron Elementary, Positive)}} &
\textcolor{blue}{\textbf{\checkmark\,(Chicago, Neutral)}} \newline \textcolor{blue}{\textbf{\customcross\,(Cameron Elementary, Neutral)}} &
\textcolor{deepgreen}{\textbf{\customcross\,(Chicago, Positive)}} \newline \textcolor{red}{\textbf{\checkmark\,(Cameron Elementary, Negative)}} &
\textcolor{blue}{\textbf{\checkmark\,(Chicago, Neutral)}} \newline \textcolor{red}{\textbf{\checkmark\,(Cameron Elementary, Negative)}} \\
\hline
\includegraphics[width=1.5cm,height=1.5cm]{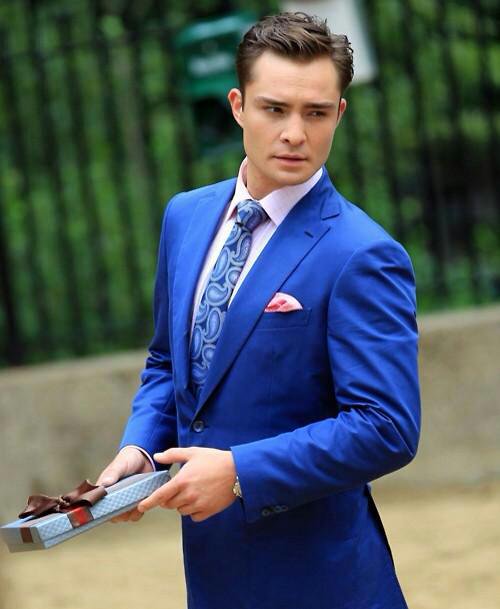} &
RT @ ltsChuckBass : Chuck Bass is everything \#MCM &
\textcolor{deepgreen}{\textbf{(Chuck Bass, Positive)}} \newline \textcolor{blue}{\textbf{(\#MCM, Neutral)}} &
\textcolor{red}{\textbf{\customcross\,(Chuck Bass, Negative)}} \newline \textcolor{blue}{\textbf{\checkmark\,(\#MCM, Neutral)}} &
\textcolor{blue}{\textbf{\customcross\,(Chuck Bass, Neutral)}} \newline \textcolor{blue}{\textbf{\checkmark\,(\#MCM, Neutral)}} &
\textcolor{deepgreen}{\textbf{\checkmark\,(Chuck Bass, Positive)}} \newline \textcolor{deepgreen}{\textbf{\customcross\,(\#MCM, Positive)}} &
\textcolor{deepgreen}{\textbf{\checkmark\,(Chuck Bass, Positive)}} \newline \textcolor{blue}{\textbf{\checkmark\,(\#MCM, Neutral)}} \\
\hline
\includegraphics[width=1.5cm,height=1.5cm]{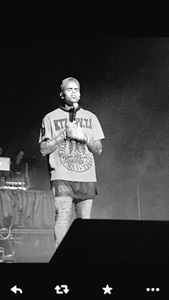} &
Why Chris Brown and Beyonce look like they tryna lead Praise and Worship? &
\textcolor{red}{\textbf{(Chris Brown, Negative)}} \newline \textcolor{red}{\textbf{(Beyonce, Negative)}} &
\textcolor{red}{\textbf{\checkmark\,(Chris Brown, Negative)}} \newline \textcolor{deepgreen}{\textbf{\customcross\,(Beyonce, Positive)}} &
\textcolor{deepgreen}{\textbf{\customcross\,(Chris Brown, Positive)}} \newline \textcolor{red}{\textbf{\checkmark\,(Beyonce, Negative)}} &
\textcolor{blue}{\textbf{\customcross\,(Chris Brown, Neutral)}} \newline \textcolor{red}{\textbf{\checkmark\,(Beyonce, Negative)}} &
\textcolor{red}{\textbf{\checkmark\,(Chris Brown, Negative)}} \newline \textcolor{red}{\textbf{\checkmark\,(Beyonce, Negative)}} \\
\hline
\includegraphics[width=1.5cm,height=1.5cm]{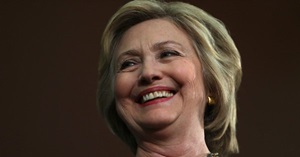} &
Donald Trump is still obsessed with Hillary Clinton’s laugh: &
\textcolor{blue}{\textbf{(Donald Trump, Neutral)}} \newline \textcolor{red}{\textbf{(Hillary Clinton, Negative)}} &
\textcolor{deepgreen}{\textbf{\customcross\,(Donald Trump, Positive)}} \newline \textcolor{red}{\textbf{\checkmark\,(Hillary Clinton, Negative)}} &
\textcolor{blue}{\textbf{\checkmark\,(Donald Trump, Neutral)}} \newline \textcolor{red}{\textbf{\checkmark\,(Hillary Clinton, Negative)}} &
\textcolor{blue}{\textbf{\checkmark\,(Donald Trump, Neutral)}} \newline \textcolor{red}{\textbf{\checkmark\,(Hillary Clinton, Negative)}} &
\textcolor{blue}{\textbf{\checkmark\,(Donald Trump, Neutral)}} \newline \textcolor{red}{\textbf{\checkmark\,(Hillary Clinton, Negative)}} \\
\hline
\end{tabular}%
}
\end{table}

Table~\ref{tab:table5} compares ground-truth sentiments with predictions from \textbf{TMFN}, \textbf{AoM}, \textbf{DPQSA}, and \textbf{AdaptiSent}, highlighting error patterns and demonstrating how our method more robustly isolates true sentiment signals. As shown, \textbf{TMFN}~\cite{wang2024tmfn} makes four errors—mislabeling \textit{Cameron Elementary}, \textit{Chuck Bass}, \textit{Beyonce}, and \textit{Donald Trump}—while \textbf{AoM}~\cite{zhou2023aom} reduces this to three by correctly identifying \textit{Trump} and \textit{Clinton} but misclassifying the rest. \textbf{DPQSA}~\cite{peng2024novel} also makes three errors, misreading \textit{Chicago}, \textit{\#MCM}, and \textit{Chris Brown}. In contrast, \textbf{AdaptiSent} achieves perfect agreement, aided by aspect-aware captioning and context-based masking.

\section{Conclusion \& Future Work}

\textbf{AdaptiSent} proposes an adaptive cross-modal attention mechanism that learns instance-specific weights for textual and visual cues, allowing finer inter-modal control. It excels over existing methods, especially in managing complex inter-modal dynamics.  Its dynamic weighting mitigates modality noise, and regularization ensures cross-modal alignment. The model’s regularization term ensures embedding alignment across modalities, improving generalization on out‐of-domain samples. Future work includes lightweight attention designs, handling misaligned inputs, and scaling to noisier datasets for real-world applicability.

We envision enhancing AdaptiSent with \textit{sentiment reasoning} capabilities, as a systems approach to \textit{neuro-symbolic integration}. e.g., integrating commonsense knowledge graphs and ontologies (e.g., SenticNet, ConceptNet) to aid interpretability and contextual grounding of sentiment predictions. Symbolic cognitive ``theory of mind" models, contrastive reasoning frameworks, and counterfactual sentiment analysis, could be effective in reasoning over complex affective phenomena like sarcasm, deception, irony, and higher-order sentiment reasoning.


\end{document}